\documentclass[letterpaper, 10 pt, conference]{ieeeconf}  % Comment this line out
\IEEEoverridecommandlockouts                              % This command is only
\usepackage{graphics} % for pdf, bitmapped graphics files
\usepackage{epsfig} % for postscript graphics files
\usepackage{mathptmx} % assumes new font selection scheme installed
\usepackage{times} % assumes new font selection scheme installed
\usepackage{amsmath} % assumes amsmath package installed
\usepackage{amssymb}  % assumes amsmath package installed
\usepackage[belowskip=-5pt,aboveskip=2pt]{caption}
\usepackage{array}
\usepackage[tight,footnotesize]{subfigure}

\title{\LARGE \bf
Fusion of Monocular Vision and Radio-based Ranging for Global Scale Estimation and Drift Mitigation
}

\author{Young-Hee Lee$^{1}$, Chen Zhu$^{2}$, Gabriele Giorgi$^{2}$ and Christoph G\"{u}nther$^{3}$% <-this % stops a space
\thanks{$^{1}$Young-Hee Lee is a research assistant at the Institute for Communications and Navigation, Technical University of Munich, Munich, Germany
{\tt\small younghee.lee@tum.de}}%
\thanks{$^{2}$Chen Zhu and Gabriele Giorgi are researchers at the Institute of Communications and Navigation, German Aerospace Center, Oberpfaffenhofen, Germany
{\tt\small chen.zhu@dlr.de}, \hspace{0.5em} {\tt\small gabriele.giorgi@dlr.de}}%%
\thanks{$^{3}$Christoph G\"{u}nther is the head of the Institute of Communications and Navigation of the German Aerospace Center and the Institute for Communications and Navigation at the Technical University of Munich
{\tt\small christoph.guenther@dlr.de}}%
\thanks{This work was conducted for the project VaMEx-CoSMiC that is supported by the Federal Ministry for Economic Affairs and Energy (on the basis of a decision by the German Bundestag, grant 50NA1521) and administered by DLR Space Administration.}
}

\newcommand{\stateVec}[3]{ {\scriptscriptstyle {}_{\mathrm{#1}}}{\displaystyle \textbf{x}}_{ \scriptscriptstyle \mathrm{#2} }^{\scriptstyle #3} }
\newcommand{\meaVec}[2]{ {\displaystyle \textbf{z}}_{ \scriptstyle \mathrm{#1} }^{\scriptstyle #2} }

\newcommand{\RMat}[3]{ {\displaystyle \mathbf{R}}_{\scriptscriptstyle \mathrm{#1#2}}^{\scriptstyle #3} }
\newcommand{\pVec}[3]{ {\scriptscriptstyle {}_{\mathrm{#1}}}{\displaystyle \textbf{p}}_{ \scriptscriptstyle \mathrm{#2} }^{\scriptstyle #3} }
\newcommand{\zVec}[3]{ {\scriptscriptstyle {}_{\mathrm{#1}}}{\displaystyle \textbf{z}}_{ \scriptstyle \mathrm{#2} }^{\scriptstyle #3} }
\newcommand{\vVec}[4]{ {\scriptscriptstyle {}_{\mathrm{#2}}}{\displaystyle \textbf{#1}}_{ \scriptscriptstyle \mathrm{#3} }^{\scriptstyle #4} } 

\begin{document}

\maketitle
\thispagestyle{empty}
\pagestyle{empty}

%%%%%%%%%%%%%%%%%%%%%%%%%%%%%%%%%%%%%%%%%%%%%%%%%%%%%%%%%%%%%%%%%%%%%%%%%%%%%%%%
\begin{abstract}
Monocular vision-based Simultaneous Localization and Mapping (SLAM) is used for various purposes due to its advantages in cost, simple setup, as well as availability in the environments where navigation with satellites is not effective. 
However, camera motion and map points can be estimated only up to a global scale factor with monocular vision. Moreover, estimation error accumulates over time without bound, if the camera cannot detect the previously observed map points for closing a loop. 
We propose an innovative approach to estimate a global scale factor and reduce drifts in monocular vision-based localization with an additional single ranging link. Our method can be easily integrated with the back-end of monocular visual SLAM methods. 
We demonstrate our algorithm with real datasets collected on a rover, and show the evaluation results. 
\end{abstract}

%%%%%%%%%%%%%%%%%%%%%%%%%%%%%%%%%%%%%%%%%%%%%%%%%%%%%%%%%%%%%%%%%%%%%%%%%%%%%%%%
\section{INTRODUCTION}
\label{sec:intro} 
Simultaneous Localization and Mapping (SLAM) with a single monocular camera is implemented for various applications for estimating camera motion and map points with a simple cost- and weight-effective setup.
However, only relative scales between estimates can be recovered with monocular vision-based approaches, since no real scale information is available. Even though a calibrated stereo rig with a known baseline length can be a simple solution to resolve the global scale ambiguity, it is not an efficient tool for mapping distant environments, since the uncertainty in the depth estimation increases quadratically with the distances of scenes.
Furthermore, the estimation error in monocular visual odometry accumulates over time without bound. The loop closing technique \cite{strasdat2010scale} is an efficient tool to correct the drifts. However, a camera has to observe the same scenes to close a loop, which becomes a significant constraint in mission design. In addition, it requires additional computational loads and data storage. Moreover, it is hard to verify the quality of loop detection. 

% Our approach
We propose a fusion of monocular visual odometry and sparse radio-based ranging to resolve global scale ambiguity and reduce drifts.
Our approach exploits a single ranging-anchor, hence the number of ranging measurements increases only linearly with time. In addition, a radio signal can cover large areas ($\sim500$m), and the measurement error is independent to the drifting estimates obtained with the vision-only process; the error does not accumulate over time. Moreover, it is easy to implement this approach, since various hardware systems are available for a wireless radio signal network.  
It should be noted that the visual-inertial system (VINS) represents the state-of-art in terms of performance in scale estimation and drift mitigation. Our approach can be either combined with the VINS to improve the robustness, or be an alternative tool. 

% Main contributions
The followings are the main contributions of the proposed sensor fusion approach:
\begin{enumerate}
\item The global scale ambiguity in monocular Vision-based SLAM (VSLAM) is reliably resolved with additional metric information. 
\item Drifts are reduced with a fusion of visual odometry and ranging measurements without closing a loop 
      (However, the loop closing technique can be still used). 
\item Our approach is evaluated with real datasets collected on a rover, achieving more accurate trajectory estimates compared to the state-of-the-art visual odometry techniques.
\item The proposed algorithm can be simply integrated with the back-end of the monocular VSLAM methods, without significantly increasing the complexity in the systems.
\end{enumerate}

% Paper structure 
We discuss related works in Section \ref{sec:rel_works}, focusing on the relation with VINS techniques. Then, we clarify our system setup and the radio-based ranging model in Section \ref{sec:system_model}, followed by the algorithm overview in Section \ref{sec:algo_overview}. In Section \ref{sec:global_scale_est}, an initial global scale estimation method using ranging measurements is proposed. Then, our graph-based sensor fusion method is presented in Section \ref{sec:global_map_refinement}. In Section \ref{sec:experiments}, we present the evaluation results of the proposed algorithm with real datasets obtained with a rover system. 

%\addtolength{\textheight}{-0cm}  % This command serves to balance the column lengths
                                  % sure that you do not shorten the textheight too much.p

%%%%%%%%%%%%%%%%%%%%%%%%%%%%%%%%%%%%%%%%%%%%%%%%%%%%%%%%%%%%%%%%%%%%%%%%%%%%%%%%
\section{Related Works}
\label{sec:rel_works}
\subsection{Monocular VSLAM}
A VSLAM system consists of two processes: the front-end to obtain geometric information with visual observations, and the back-end to estimate camera motions and construct a map database using the acquired information in the front-end. 

In feature-based VSLAM \cite{PTAM} \cite{ORB-SLAM}, feature points are extracted in the front-end process, and used for further estimation. Other approaches are available; for example, pixel intensity can be directly used to detect motion between frames under the assumption of constant light condition \cite{LSD-SLAM}.
The latter method is more efficient compared to feature-based algorithms in terms of computational complexity, since pixels are directly used without the additional processing that is needed to extract feature information. 
%However, the constant light assumption is not practical for many applications. 
% Gao et al. propose a hybrid solution: Direct Sparse Odometry with Loop Closure (LDSO) \cite{LDSO}, to improve the robustness of the direct methods against lighting changes.

In the back-end of the VSLAM, camera motion and map points are estimated with the geometric information provided from the front-end. 
A Bayesian filter, such as an Extended Kalman Filter (EKF) is an efficient method for real-time applications \cite{MonoSLAM}. 
Keyframe-based graph optimization is applied to carry out the estimation processes as well, such as the batch least-square estimation (bundle adjustment) \cite{PTAM}. To achieve real-time processing, state-of-art methods marginalize the graph, increasing the numerical efficiency.
Kaess et al. propose procedural factor graph marginalization to achieve a trade-off between speed and accuracy in the VSLAM \cite{iSAM}, \cite{iSAM2}. 

Our work focuses on improving the back-end with radio-based ranging measurements. We use the graph-based framework, treating the ranging measurements as additional edges in the factor graph. 

\subsection{Visual-ranging system}
Global Navigation Satellite System (GNSS) is one of the generally used positioning system, based on pseudo-range measurements derived from broadcast satellite radio signals. 
However, the signal from satellites is not available in many cases due to obstructions or lack of infrastructures (e.g. urban canyons, indoors, or other planets). In such environments, ranging can be provided with other systems, such as signal of opportunities (SOP) \cite{SOP}. 
Morales et al. \cite{morales_SOP_inertial} propose navigation with a fusion of a cellular network and inertial sensors. 

Tabibiazar and Basir \cite{MonoCellular} propose an approach using a monocular camera and range measurements from a cellular network, using a loosely-coupled method for localization. It requires a minimum of three ranging links, which are not available in many circumstances (e.g. mobile networks). Our method only needs a single ranging link from a radio beacon or any other SOP sources.

On-board lidar or radar sensors can provide ranging measurements as well. Zhang et al. \cite{Zhang_visual_lidar_odometry}~\cite{Zhang_depth_VO} propose a fusion of monocular vision and lidar-based ranging. 
Unlike the aforementioned methods, which use dense measurements obtained with an on-board sensor, we exploit a system that provides sparse distance information from a fixed anchor point which is independent from the camera motion estimation. Therefore, the measurement error is not accumulating over time. Furthermore, our method does not considerably increase the system complexity. 

\subsection{Visual-inertial system (VINS)}
N\"{u}tzi et al. \cite{MonoIMU_scale_ETH} propose scale estimation with inertial measurements using a tightly-coupled sensor fusion.
Since the VINS performs very efficiently, the majority of the state-of-art monocular VSLAM algorithms provides the VINS, integrated with their vision-only solution \cite{ORBSLAM_INS} \cite{LSD_INS} \cite{VIDSO}. 

Despite of its verified performance, the VINS approaches have drawbacks. First, the global consistency of estimation cannot be refined since on-board inertial sensors are used in the system. 
Furthermore, inertial bias calibration is coupled with the VSLAM estimation, causing difficulty in error analysis and system monitoring.  

We use a ranging system that provides absolute distance measurements in which errors are not coupled with the dead-reckoning vision-only estimation, hence errors can be analyzed easily. In addition, the calibration process is simpler, compared to the VINS. Furthermore, our method can be easily integrated with the VINS.

%%%%%%%%%%%%%%%%%%%%%%%%%%%%%%%%%%%%%%%%%%%%%%%%%%%%%%%%%%%%%%%%%%%%%%%%%%%%%%%%
\section{System Model}
\label{sec:system_model}
In this section, we introduce our system setup, including the frame definition and notations that are used throughout this paper, followed by the measurement model of radio-based ranging (Two-way Time-of-Flight). 

% [TODO] Location
\begin{figure}[t]
\centerline{\includegraphics[scale=0.50]{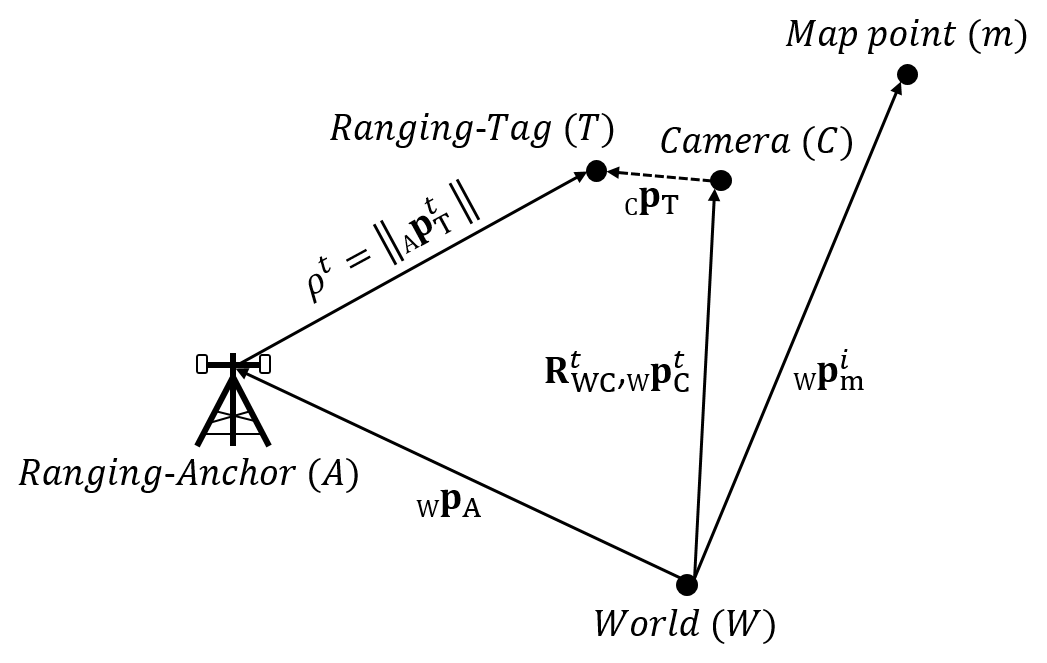}}
\caption{System setup and the relative geometry between the frames}
\label{fig:SystemSetup}
\end{figure}

% [TODO] Location
\begin{figure*}[tp]
\centerline{\includegraphics[scale=0.45]{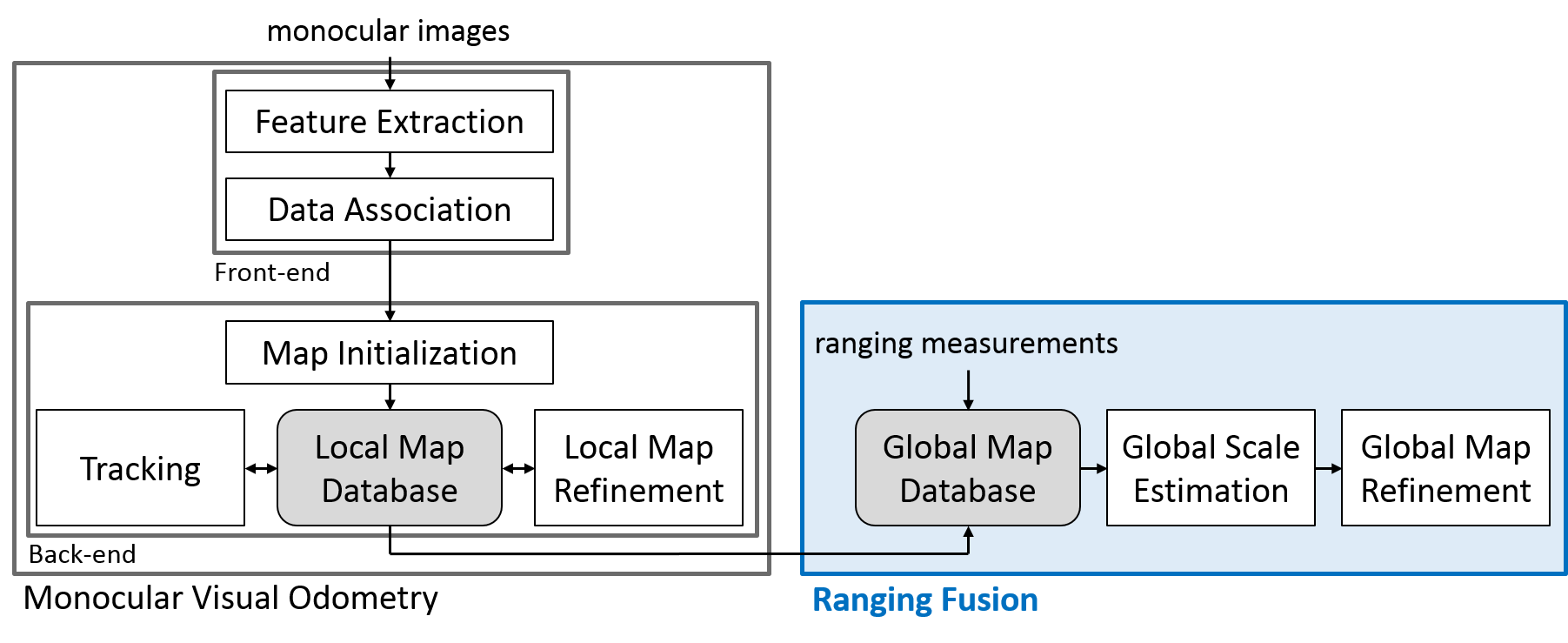}}
\caption{The flowchart of the proposed algorithm}
\vspace{5mm}
\label{fig:FlowChart}
\end{figure*}

% [TODO] Location
\begin{figure*}[t]
\centerline{\includegraphics[scale=0.50]{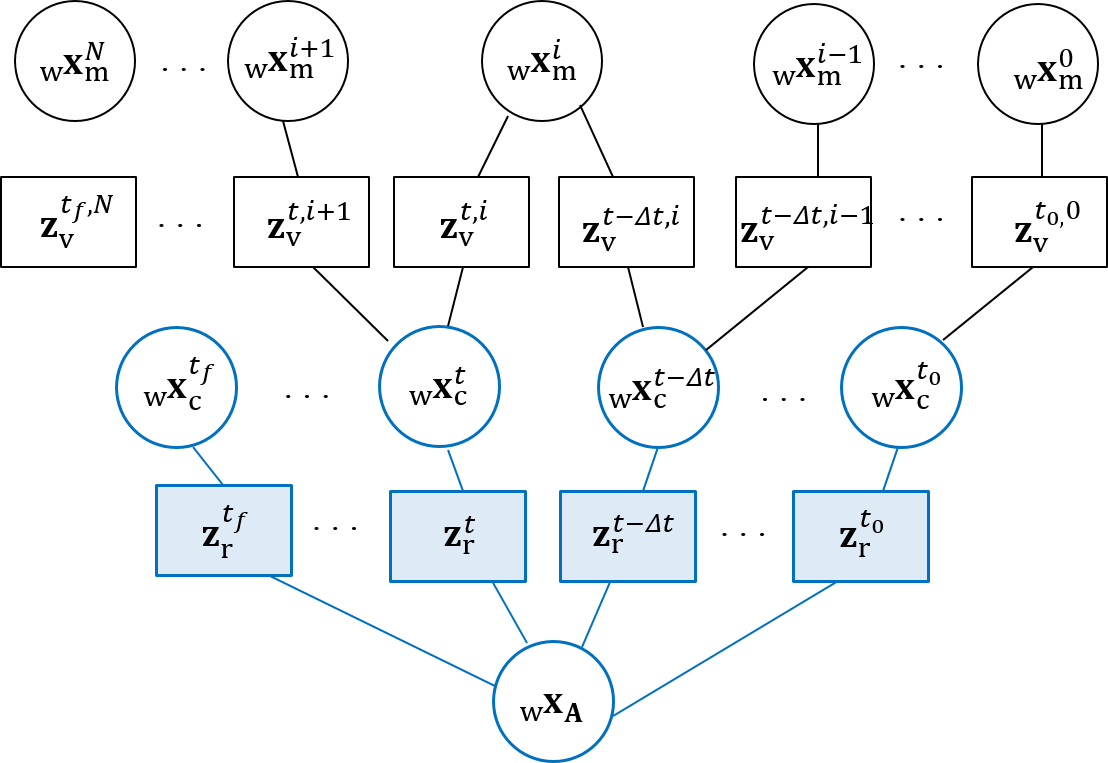}}
\caption{The graph representation of the global map database. $\stateVec{}{×}{×}$ denotes the state variables in the graph: the camera pose with respect to the \textit{World frame} at $t$ ($\stateVec{W}{C}{t}$), and the $i$-th map point position with respect to the \textit{World frame} ($\stateVec{W}{m}{i}$). Constraints are denoted with $\meaVec{}{×}$: the re-projection error of the $i$-th map point in the camera frame at $t$ ($\meaVec{v}{t,i}$), and the distance error at $t$ ($\meaVec{r}{t}$) }
\label{fig:Graph}
\end{figure*}

\subsection{Frame definition}
Fig.~\ref{fig:SystemSetup} shows our system setup and the relative geometry between frames. We mount a camera on a rover to log monocular image data. The \textit{Camera frame} is attached to the current camera body frame (\textit{C}), and the \textit{World frame} (\textit{W}) is defined by the initial \textit{Camera frame} when monocular visual odometry successfully initializes map points (\textit{m}). 
For the ranging system, we mount a ranging-tag (\textit{T}) on the rover with the camera, and install a ranging-anchor (\textit{A}) at a fixed position. These ranging nodes provide the relative distance measurement $\rho^t$ between the rover and the anchor point. The position vector of the anchor with respect to the \textit{World frame} $\pVec{W}{A}{×}$ can be estimated by trilaterating multiple distance measurements collected at the initial phase of a mission \cite{trilateration2005revisiting}~\cite{ION2017}. 
An orthonormal rotation matrix $\mathbf{R}$ describes a 3D camera rotation, and a 3D spatial position vector is denoted with $\mathbf{p}$. We use subscripts for the frame information and superscripts for the time stamps of the camera poses or map point numbers. For example, the rotation matrix from the \textit{Camera frame} to the \textit{World frame} at time $t$ is $\RMat{W}{C}{t}$, and the 3D position of the \textit{i-th} map point with respect to the \textit{World frame} is $\pVec{W}{m}{i}$.

\subsection{Radio-based ranging}
\label{subsec:radio_ranging}
We use the Two-way Time-of-Flight (TW-ToF) ranging technique \cite{mazomenos2011two} to obtain the relative distance measurements between two wireless sensor nodes (ranging-tag and ranging-anchor in Fig.~\ref{fig:SystemSetup}). First, one of the nodes (node A) transmits a signal to the other (node B), and the node B sends the signal back to the node A. Node A collects the round-trip-time of the signal multiple times, and computes an average of the whole set. Then, we can compute the relative distance between two nodes using the averaged time $\hat{t}_{\scriptscriptstyle ToF}$:

\begin{equation}
\label{eq:average_ToF}
 d = \frac{c}{2} \left( \hat{t}_{\scriptscriptstyle ToF} - t_{off} \right)
\end{equation}

where $c$ is the speed of light, and $t_{off}$ is a time offset including all delays in the system. 
Since the TW-ToF ranging system uses the signal round-trip time, it does not require time synchronization of the nodes with high precision which is problematic for other ranging systems using wireless sensor networks.

%%%%%%%%%%%%%%%%%%%%%%%%%%%%%%%%%%%%%%%%%%%%%%%%%%%%%%%%%%%%%%%%%%%%%%%%%%%%%%%%
\section{Algorithm Overview}
\label{sec:algo_overview}
The flowchart of the proposed algorithm is given in Fig.~\ref{fig:FlowChart}. 
We use the state-of-art feature-based algorithm ORB-SLAM \cite{ORB-SLAM} without closing loops to perform monocular visual odometry; any other algorithms can be applied as well. 
First, we extract feature information (the features' positions and corresponding descriptors) from the received images and store it in a map database. Then, we try to find feature matches between the images taken at different locations (the front-end of visual odometry). 
If the number of matched features is larger than a given threshold, we initialize map points by triangulation. Once the map points are successfully initialized, we start to track the current features in the database to iteratively estimate camera poses. Meanwhile, a local map (a subset of the global map that only includes the camera poses and map points correlated with the current ones) is continuously optimized such that the summation of the squared re-projection errors is minimized (the back-end of visual odometry). 
This is an effective approach for real-time localization since it does not require large data processing. Local map refinement is conducted in a separated processing unit, hence real-time tracking is not hindered. 
However, a global scale of the map cannot be estimated with monocular vision. Moreover, estimation errors are accumulating over time. 

Therefore, we propose to fuse the distance measurements between a rover and an anchor point with monocular vision. First, we construct a graph-based global map database including the ranging measurements as shown in Fig.~\ref{fig:Graph}. The variables denoted with circles are the state variables in the graph: the keyframe poses $\vVec{x}{W}{C}{t}$ and map point positions $\vVec{x}{W}{M}{i}$. Constraints are denoted with the squares on the edges between the corresponding nodes; re-projection errors (visual constraints) $\zVec{}{v}{t,i}$ and distance errors (ranging constraints) $\zVec{}{r}{t}$. After constructing the global map, we estimate an initial guess of the global scale using the monocular visual odometry results and ranging measurements. Then, we scale the keyframe and map point positions with this value. Lastly, we perform a global least-squares estimation (global LSE) with the Levenberg-Marquardt algorithm aiming at minimizing the squared weighted sum of both re-projection and ranging residuals. This is described in the following section.

%%%%%%%%%%%%%%%%%%%%%%%%%%%%%%%%%%%%%%%%%%%%%%%%%%%%%%%%%%%%%%%%%%%%%%%%%%%%%%%%
\section{Global Scale Estimation}
\label{sec:global_scale_est}
Since no metric information is available in a monocular vision-only method, the scales of the camera translation and map points are randomly determined at the initialization stage (e.g. normalization). After the initialization, the system generates the trajectory and new map points with the scales that are relative to the initialized value; if no error is introduced in the visual odometry process, we can recover the up-to-scale estimates by multiplying with a true global scale $\alpha$. 

To resolve this global scale ambiguity, we propose to use an additional ranging system which provides the distance measurements between a rover and an anchor $\hat{\rho}^t$. The ranging measurement model is

\begin{equation}
\label{eq:DistMeaAndScaleRelation} 
  \hat{\rho}^t = || \pVec{A}{T}{t} || = || -\pVec{W}{A}{×} + \alpha\pVec{W}{C}{t} + \RMat{W}{C}{t}\pVec{C}{T}{×}||
\end{equation}

For each value of $\rho$, two solutions for the global scale can be found analytically, if no error is modeled for both odometry and ranging:

\begin{equation}
\label{eq:TwoAlphaCandidates}
  \begin{array}{c}
   \alpha_0, \alpha_1 = -\mathrm{B}/\mathrm{A} \pm \sqrt{ \left( \mathrm{B}/\mathrm{A} \right)^2 - \mathrm{C}/\mathrm{A} } \\
   \text{where   } 
      \begin{cases}
        \mathrm{A} = || \pVec{W}{C}{t} ||^2 \\
        \mathrm{B} = {\pVec{W}{C}{t}}^T  ( \RMat{W}{C}{t}\pVec{C}{T}{×} - \pVec{W}{A}{×} ) \\
        \mathrm{C} = || \RMat{W}{C}{t}\pVec{C}{T}{×} - \pVec{W}{A}{×} ||^2 - ( \hat{\rho}^t )^2
      \end{cases}
  \end{array}
\end{equation}

\begin{figure}[t]
\centerline{\includegraphics[scale=0.50]{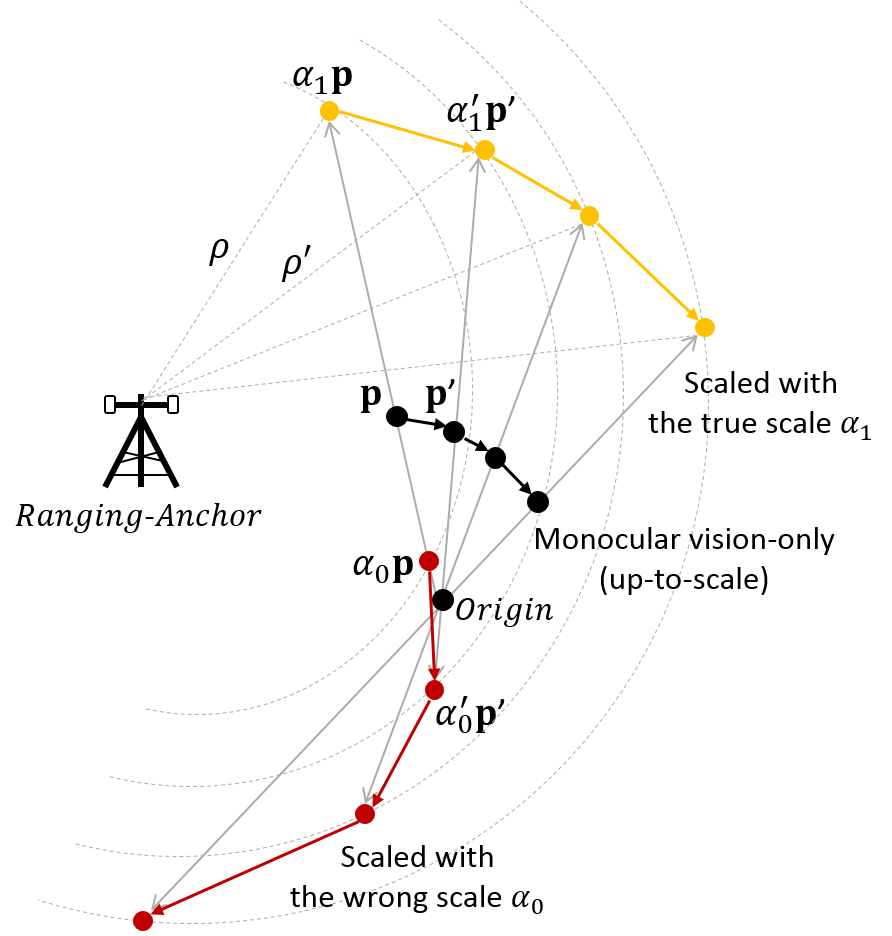}}
\caption{
When there is no error in the distance measurements $\rho$ and the trajectory generated with a monocular vision-only method (black), the true scale $\alpha_1$ is constant, enabling the recovery of the true trajectory (orange). The relative scales and direction in the up-to-scale trajectory are maintained in the true trajectory. The wrong candidate $\alpha_0$ is a different value for each ranging measurement, and the relative geometry is not preserved in the trajectory scaled with the wrong value (dark red)}
\label{fig:TwoAlphaCandidates}
\end{figure}

Fig.~\ref{fig:TwoAlphaCandidates} shows the physical meaning of the solutions. A single ranging measurement $\rho$ does not allow to discriminate between two candidates ($\alpha_0, \alpha_1$). 
When another relative distance $\rho'$ is available, we can compute an additional duplet. 
If there is no error in visual odometry and ranging measurements, one of the factors remains constant for each ranging: this value represents the recovered global scale ($\alpha_1$ in Fig.~\ref{fig:TwoAlphaCandidates}). Moreover, the relative geometry of the estimates generated with a monocular vision-only method (black in Fig.~\ref{fig:TwoAlphaCandidates}) is maintained in the trajectory scaled with the true global scale (orange in Fig.~\ref{fig:TwoAlphaCandidates}). 
However, the other candidate is a different value at each time. In addition, the relative scales and direction of the up-to-scale trajectory are not maintained in the trajectory scaled with the wrong value (dark red in Fig.~\ref{fig:TwoAlphaCandidates}). 

In the real world, errors exist in the vision-only estimates and ranging. Therefore, we collect a number of estimated duplets over time, and select the mean value of $\alpha_i$ characterized by the smallest standard deviation. 

% [TODO] Location

%%%%%%%%%%%%%%%%%%%%%%%%%%%%%%%%%%%%%%%%%%%%%%%%%%%%%%%%%%%%%%%%%%%%%%%%%%%%%%%%
\section{Global Map Refinement}
\label{sec:global_map_refinement}
After the global scale initialization, all keyframes and map point positions are accordingly scaled. Then, we carry out the global least-squares estimation such that the summation of the squared re-projection and distance errors in the global map database is minimized. 
Ranging measurements provide absolute metric information, which is not coupled with the parameters estimated with the vision-based method, enabling mitigation of drifts in the dead-reckoning process.

%%%%%%%%%%%%%%%%%%%%%%%%%%%%%%%%%%%%%%%%%%%%%%%%%%%%%%%%%%%%%%%%%%%%%%%%%%%%%%%%
\section{Experimental Tests}
\label{sec:experiments}
We conducted a number of tests in an outdoor environment to collect datasets; no public available dataset provides images and ranging simultaneously. The rover system developed by the Institute of Communications and Navigation, German Aerospace Center (DLR-KN) was used as a platform. 

In this section, we present the test results of the proposed algorithm, conducted with the collected real dataset. 

First, we show the initial global scale estimation results. Then, we discuss the trajectory and map points estimated with the proposed approach. Moreover, the performance of trajectory estimation is compared quantitatively with those provided by other algorithms, such as the trajectory estimated with monocular visual odometry scaled with the ground truth and the initial global scale guess, and stereo visual odometry.

\subsection{Setup}
\begin{figure}[t]
\centerline{\includegraphics[scale=0.34]{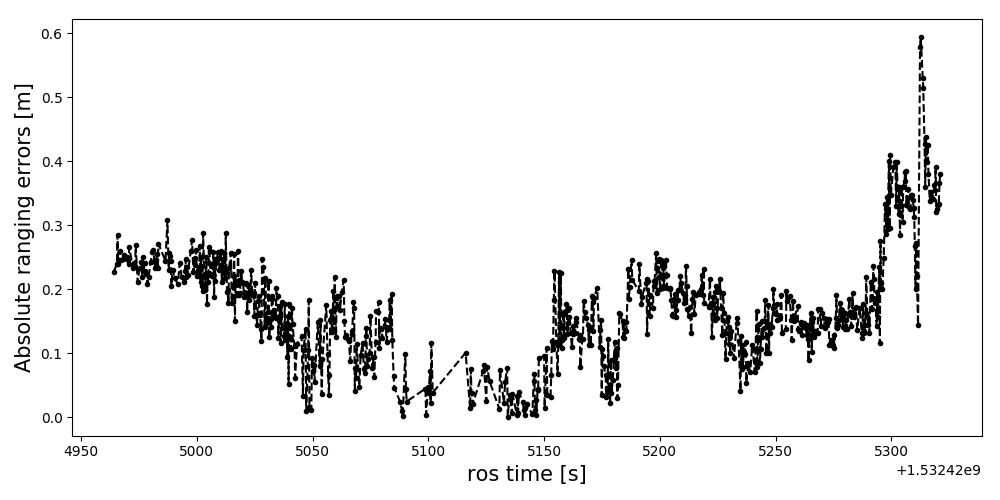}}
\caption{Absolute differences between the ranging measurements and the true distances computed with the ground truth trajectory in [m]}
\label{fig:RangingMea_Err}
\end{figure}

We used the left camera of the stereo camera Bumblebee2 (FLIR) to obtain monocular images ($1024\times768$, $30$ fps), and two ultra-wideband (UWB) wireless sensors (Decawave) to collect distance interruptions. 
The UWB sensors provide the distance measurements between the rover and an anchor point with an accuracy of $\pm 10$ cm under ideal circumstances, such as no multipath or Line-of-Sight interruptions. 
These sensors were integrated them in the rover system developed by the Institute of Communications and Navigation, German Aerospace Center (DLR-KN); we mounted the camera and a UWB sensor (ranging-tag) on a rover, and installed another UWB sensor (ranging-anchor) at the base station. 
The measurement campaign was conducted on the soccer field located in the German Aerospace Center, Oberpfaffenhofen, collecting images and ranging, as well as the ground truth trajectory with a GNSS- and Inertial Measurement Unit (IMU)-based Real Time Kinetic (RTK) system for a quantitative evaluation. Fig.~\ref{fig:ImgWithFeatures} is an example image taken with features extracted in the frame, and Fig.~\ref{fig:RangingMea_Err} shows the absolute differences between the ranging measurements and the true values computed with the ground truth trajectory. 

\subsection{Results}
\begin{figure*}[tp]
  \centerline
  {
    \subfigure[The trajectory estimates multiplied with two global scale candidates $\alpha_0$ (dark red) and $\alpha_1$ (orange) in Table~\ref{table:TwoAlphaResults}]
    {\includegraphics[scale=0.46]{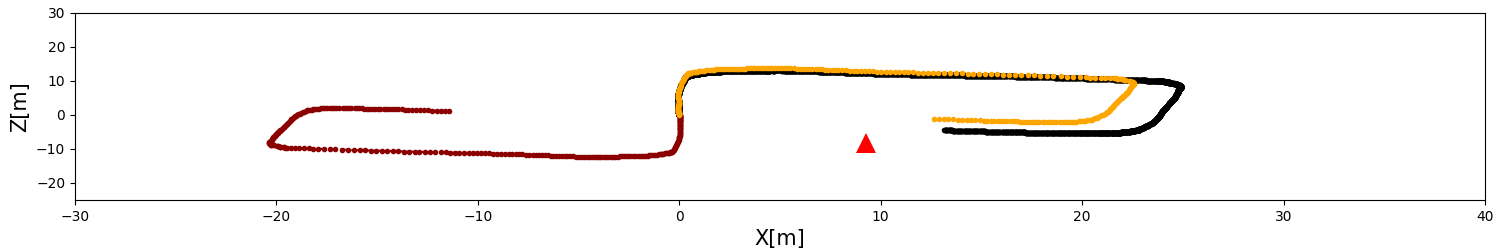}
    \label{fig:Traj_MonoVO_with_Alphas}}
  }
  \centerline
  {
    \subfigure[The trajectory obtained with the proposed algorithm (green), and the one scaled with the true scales computed with the ground truth trajectory (blue)]
    {\includegraphics[scale=0.44]{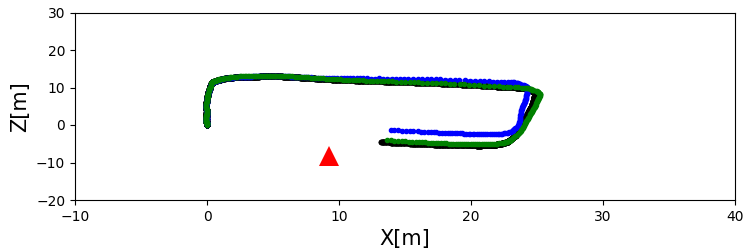}
    \label{fig:Traj_MonoVO_with_GS}}
    ~~~
    \subfigure[The trajectory obtained with the proposed algorithm (green), and the one estimated with stereo visual odometry (purple)]
    {\includegraphics[scale=0.44]
    {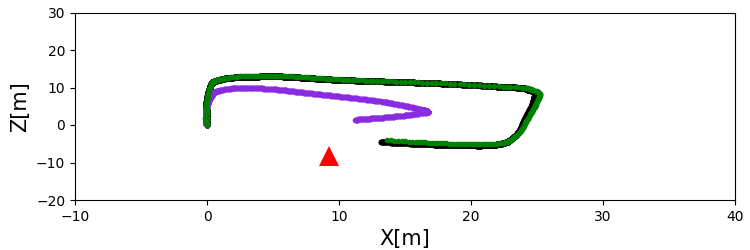}
    \label{fig:Traj_StereoVO}}
   }
  \caption
  {The trajectory estimates in the horizontal plane: the ground truth trajectory is represented as a black line, and the anchor position is denoted with a red triangle in each figure}
  \label{fig:Traj_Comparison}
\end{figure*}

Table~\ref{table:TwoAlphaResults} shows the mean and standard deviation of two global scale candidates $\alpha_0$ and $\alpha_1$. The mean value of $\alpha_1$ is chosen as an initial global scale guess since the standard deviation of $\alpha_1$ is smaller than the one of $\alpha_0$. 
Fig.~\ref{fig:Traj_MonoVO_with_Alphas} presents the trajectory scaled with $\alpha_0$ (dark red) and $\alpha_1$ (orange), as well as the ground truth trajectory (black) with respect to the \textit{World frame}. It qualitatively shows that $\alpha_1$ is a proper initial guess of the global scale compared to $\alpha_0$. 

\begin{figure*}[tp]
\centerline{\includegraphics[scale=0.47]{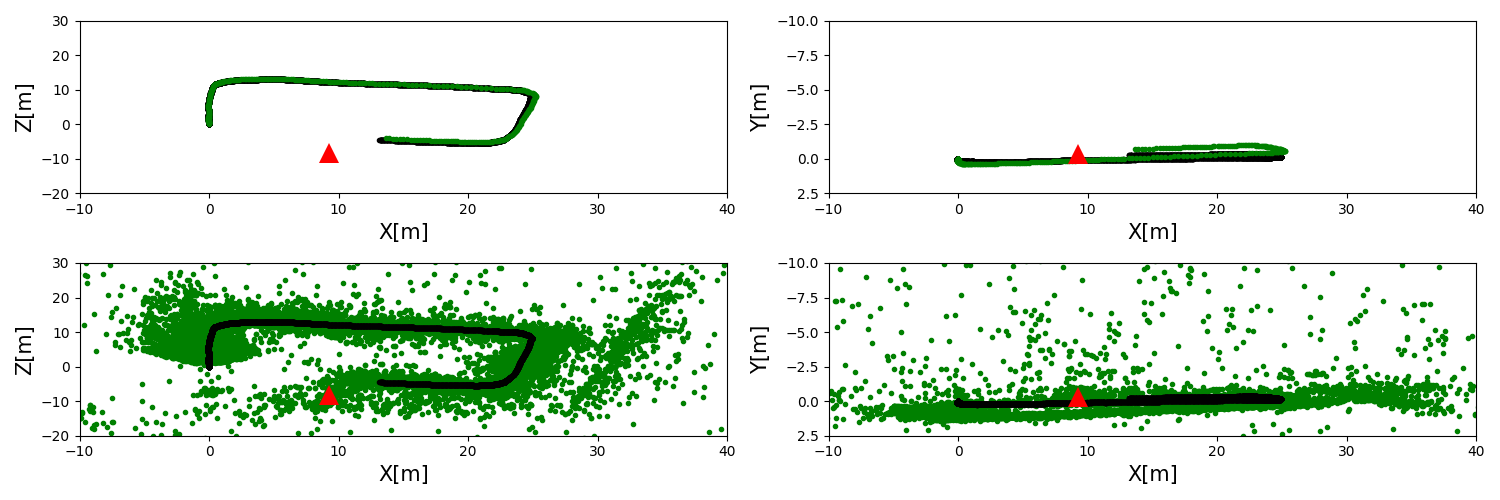}}
\caption{The trajectory (top) and map points (bottom) obtained with the proposed fusion algorithm in the horizontal (left) and vertical (right) planes}
\label{fig:TrajAndMapPointsOfOpt}
\end{figure*}

Fig.~\ref{fig:TrajAndMapPointsOfOpt} shows the trajectory and map point estimates with respect to the \textit{World frame} after conducting the proposed global least-squares estimation. 
Map point positions cannot be evaluated quantitatively since there is no ground truth, whereas the performance of the trajectory estimation can be quantified by RMSE using the ground truth obtained with the RTK system. 
As shown in Table~\ref{table:TrajRMSE}, even though we scale the monocular visual odometry trajectory with the true scales computed with the ground truth trajectory, the RMSE is relatively high. In addition, the trajectory scaled with the initial global scale guess shows considerable RMSE as well. After the global least-squares estimation, the RMSE of the trajectory estimation becomes significantly lower than the others. This result shows that our global map refinement reduces accumulated estimation errors.

\begin{figure}[t]
\centerline{\includegraphics[scale=0.27]{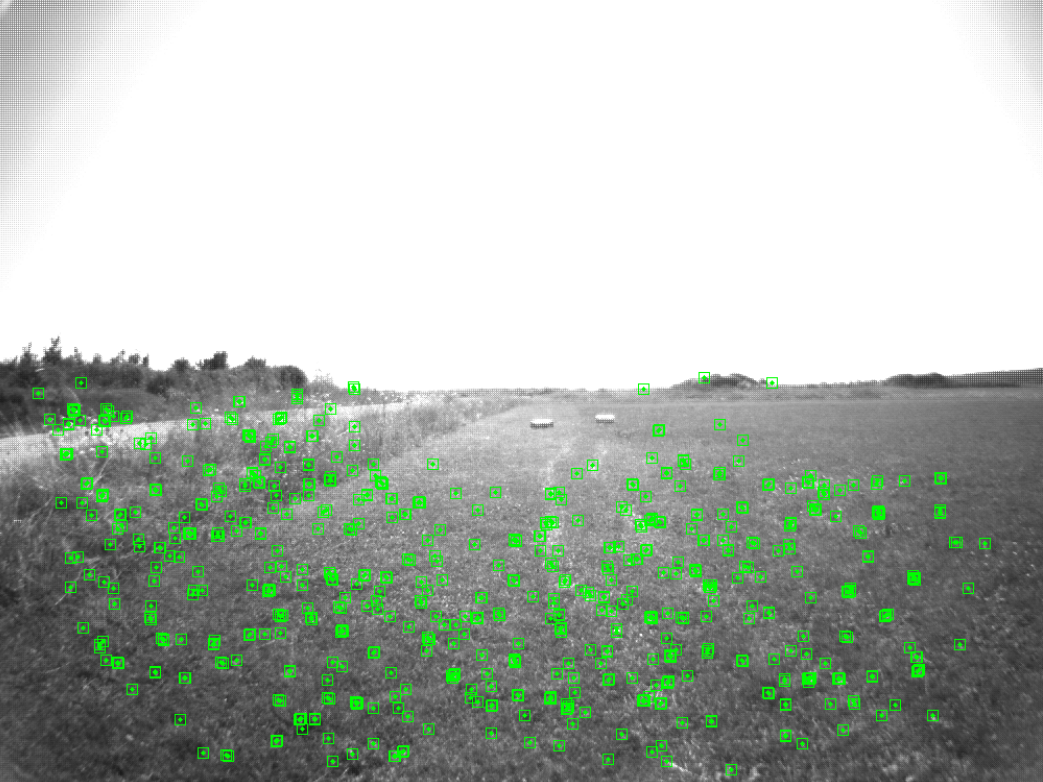}}
\caption{An example of the image with the extracted features}
\label{fig:ImgWithFeatures}
\end{figure}

Additionally, we tested stereo ORB-SLAM \cite{ORB-SLAM} without closing loops, using the real stereo images collected with the stereo camera. 
As shown in Fig.~\ref{fig:ImgWithFeatures}, the detected features are mostly in the grass field, which cannot be easily tracked between frames (low quality features). Moreover, we used a stereo camera with a short baseline length ($\sim12$ cm). Therefore, the performance is worse than the one of the proposed algorithm, as well as monocular visual odometry scaled with true scales as shown in Fig.~\ref{fig:Traj_StereoVO} and Table~\ref{table:TrajRMSE}.

% [TODO] Location

% [TODO] Location

%[TODO] Location

%[TODO] Location

% [TODO] Location
\begin{table*}[tp]
  \caption{The mean and standard deviation of two global scale candidates}
  \begin{center}
  \begin{tabular}{ ccc }
     \hline
      \textbf{} & \textbf{Mean} & \textbf{Standard deviation} \\
      \hline \hline
      $\alpha_0$ & -4.17613  & 9.39333   \\
      \hline
      $\alpha_1$ &  4.63161 & 0.213666 \\
     \hline
  \end{tabular}
  \label{table:TwoAlphaResults}
  \end{center}
\end{table*}

% [TODO] Location
\begin{table*}[t]
  \caption{The RMSE of the trajectory estimates in [m]}
  \begin{center}
  \begin{tabular}{ lc }
     \hline
      \textbf{Algorithms} & \textbf{RMSE[m]}  \\
      \hline \hline
      \textrm{MonoVO scaled with the true scales (blue in Fig.~\ref{fig:Traj_Comparison})} & 1.742    \\
      \hline
      \textrm{MonoVO scaled with the estimated global scale, without global LSE (orange in Fig.~\ref{fig:Traj_Comparison})} & 2.206   \\
      \hline
      \textrm{Trajectory estimated with the proposed approach, with global LSE (green in Fig.~\ref{fig:Traj_Comparison})} & 0.450    \\
      \hline
      \textrm{StereoVO (purple in Fig.~\ref{fig:Traj_Comparison})} & 6.833    \\
      \hline
  \end{tabular}
  \label{table:TrajRMSE}
  \end{center}
\end{table*}

%%%%%%%%%%%%%%%%%%%%%%%%%%%%%%%%%%%%%%%%%%%%%%%%%%%%%%%%%%%%%%%%%%%%%%%%%%%%%%%%
\section{CONCLUSION}
We proposed a fusion of radio-based ranging measurements and monocular vision, enabling global scale estimation and drift mitigation in monocular visual odometry with a single ranging link. We tested the algorithm with a real dataset collected on a rover, showing more accurate performance in the trajectory estimation compared to other algorithms (monocular visual odometry scaled with the true scales and the initial global scale guess, and stereo visual odometry).

The proposed method can be easily integrated within the VSLAM's back-end with a simple additional setup. In addition, our method can be an auxiliary option to the VINS (or integrated with the VINS) for resolving the scale ambiguity problem. 

As a further step, we will test the proposed algorithm on a number of different scenarios. The initial global scale estimation method will be improved as well, improving robustness against outliers in ranging measurements. Furthermore, real-time performance will be improved, implementing the algorithm on an additional processing unit (separated from real-time tracking).

%%%%%%%%%%%%%%%%%%%%%%%%%%%%%%%%%%%%%%%%%%%%%%%%%%%%%%%%%%%%%%%%%%%%%%%%%%%%%%%%
%\section*{APPENDIX}

%\section*{ACKNOWLEDGMENT}

%%%%%%%%%%%%%%%%%%%%%%%%%%%%%%%%%%%%%%%%%%%%%%%%%%%%%%%%%%%%%%%%%%%%%%%%%%%%%%%%
\bibliography{main}
\bibliographystyle{IEEEtran}

\end{document}